\documentclass[sn-mathphys-num]{sn-jnl}

\usepackage{graphicx}%
\usepackage{multirow}%
\usepackage{amsmath,amssymb,amsfonts}%
\usepackage{amsthm}%
\usepackage{mathrsfs}%
\usepackage[title]{appendix}%
\usepackage{xcolor}%
\usepackage{textcomp}%
\usepackage{manyfoot}%
\usepackage{booktabs}%
\usepackage{algorithm}%
\usepackage{algorithmicx}%
\usepackage{algpseudocode}%
\usepackage{listings}%
\usepackage{bm}%
\theoremstyle{thmstyleone}%

\theoremstyle{thmstyletwo}%

\theoremstyle{thmstylethree}%

\DeclareRobustCommand{\myref}[2]{\hyperref[#1]{\autoref{#1}#2}}

\begin{document}

\title[Article Title]{Human-Exoskeleton Interaction Portrait }


\author[1]{\fnm{Mohammad} \sur{Shushtari}}\email{smshushtari@uwaterloo.ca}

\author[2]{\fnm{Julia} \sur{Foellmer}}\email{julia.foellmer@tuhh.de}

\author*[1,3]{\fnm{Arash} \sur{Arami}}\email{arash.arami@uwaterloo.ca}

\affil*[1]{Department of Mechanical and Mechatronics Engineering, University of Waterloo, Waterloo, ON N2L 3G1, Canada.}

\affil[2]{Mechanics and Ocean Engineering Department, Hamburg University of Technology, 21071 Hamburg, Germany}

\affil[3]{Toronto Rehabilitation Institute (KITE), University Health Network, Toronto, ON M5G 2A2, Canada}


\abstract{Human-robot physical interaction contains crucial information for optimizing user experience, enhancing robot performance, and objectively assessing user adaptation. This study introduces a new method to evaluate human-robot co-adaptation in lower limb exoskeletons by analyzing muscle activity and interaction torque as a two-dimensional random variable. We introduce the Interaction Portrait (IP), which visualizes this variable's distribution in polar coordinates. We applied this metric to compare a recent torque controller (HTC) based on kinematic state feedback and a novel feedforward controller (AMTC) with online learning, proposed herein, against a time-based controller (TBC) during treadmill walking at varying speeds. 
Compared to TBC, both HTC and AMTC significantly lower users' normalized oxygen uptake, suggesting enhanced user-exoskeleton coordination. IP analysis reveals this improvement stems from two distinct co-adaptation strategies, unidentifiable by traditional muscle activity or interaction torque analyses alone. HTC encourages users to yield control to the exoskeleton, decreasing muscular effort but increasing interaction torque, as the exoskeleton compensates for user dynamics. Conversely, AMTC promotes user engagement through increased muscular effort and reduced interaction torques, aligning it more closely with rehabilitation and gait training applications. IP phase evolution provides insight into each user's interaction strategy development, showcasing IP analysis's potential in comparing and designing novel controllers to optimize human-robot interaction in wearable robots.}


\maketitle

\section{Introduction}\label{sec1}
Assistive and rehabilitation robotics are gaining increasing attention as they deliver a more substantial dose of exercise to users, enhancing their functionality and quality of life while reducing the workload of physical therapists~\cite{Dupont2021,DuschauWicke2009}. Despite recent advancements, including human-in-the-loop optimization to improve exoskeleton assistance~\cite{Bryan2021,Franks2021,Durandau2022,Poggensee2021}, these robotic systems still lack the sophistication to automatically fine-tune the level of support required for each individual user effectively~\cite{Nuckols2021,Lee2023}. This personalized touch, instinctive for physical therapists in traditional therapy sessions, remains a challenge for robots. The challenge arises because, although separate performance indicators like metabolic cost~\cite{Postol2020,Witte2020}, muscle activation~\cite{Zhu2021}, interaction forces~\cite{Kuecuektabak2024}, comfort~\cite{Eva2020}, cognitive load~\cite{Masengo2023}, and user preference~\cite{Ingraham2022} are utilized, the lack of a unified metric that fully encapsulates the nuances of human-robot physical interactions obstructs the precise adjustment and customization of lower limb exoskeleton support~\cite{PintoFernandez2020}. Therefore, this task is often delegated to adaptive controllers with implicit considerations to control the human-exoskeleton physical interaction. The control of human-exoskeleton interaction plays a key role in optimizing the user experience and performance of lower limb exoskeletons for rehabilitation as well as power augmentation applications~\cite{Slade2022}. In power augmentation scenarios, the user retains full autonomy, and the exoskeleton follows user commands directly or indirectly by interpreting their intended motion. In case of disagreement between the user and the exoskeleton, the exoskeleton must relinquish control in favor of the user~\cite{Medina2015,Martinez2018}. However, in the context of rehabilitation and assistive exoskeletons, human-exoskeleton interaction control is more challenging due to two primary factors. First, the user-performed motion is not always reliable due to musculoskeletal or motor impairments~\cite{Shushtari2021} which may undermine the quality of decoded intention solely based on user-robot physical interaction. Second, the exoskeleton should encourage the user to maximize their engagement in motion when possible and assist or correct when the user is unable to perform the motion correctly~\cite{Asl2020,Dominijanni2023}. Consequently, the exoskeleton must seamlessly transition between the leader and follower roles~\cite{Losey2018}.

\begin{figure*}[t!]
	\centering	
	\scalebox{.4}{\includegraphics[trim=40pt 0 0 0]{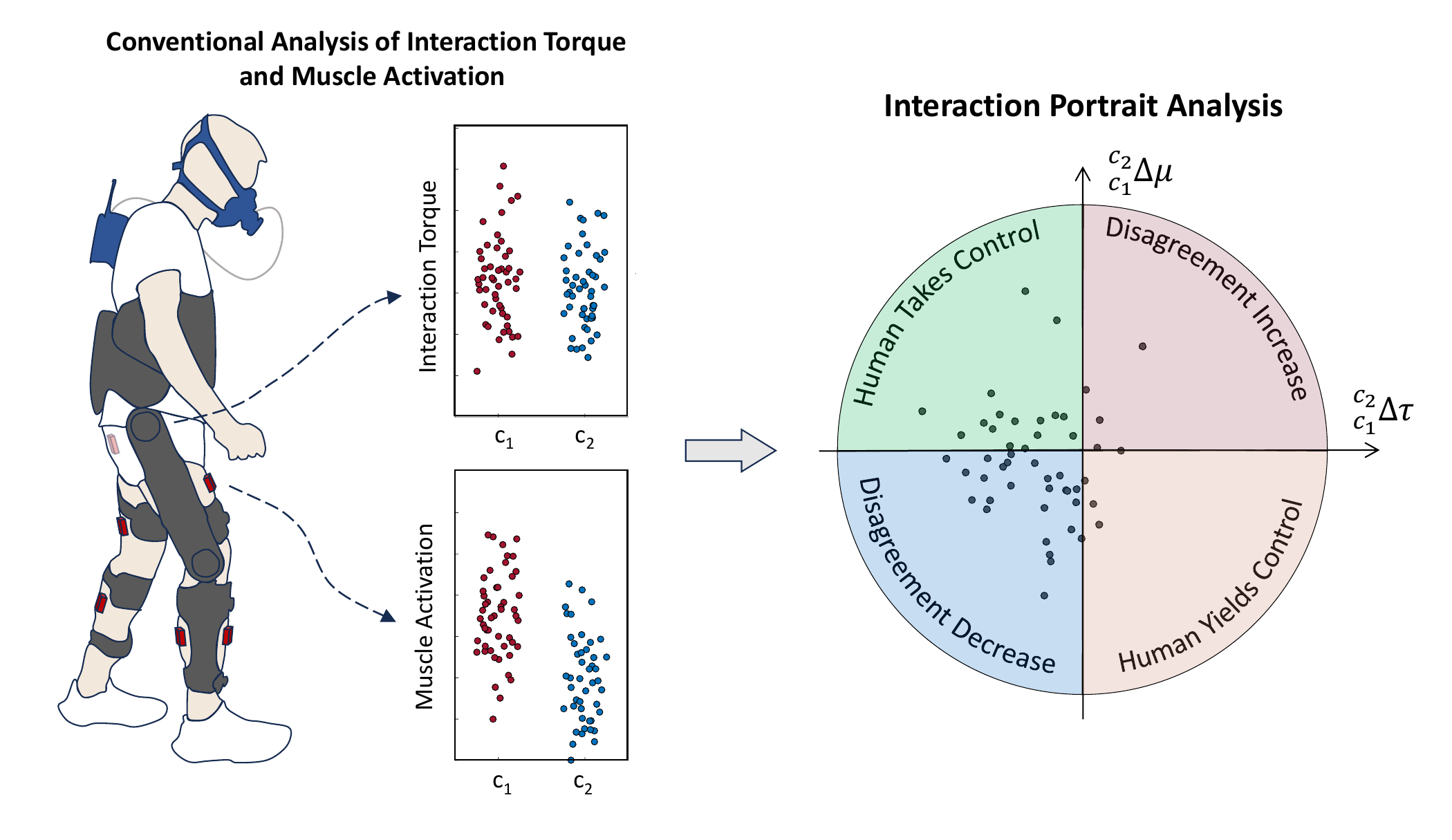}}
	\caption{Regions of Interaction Portrait (IP). Each quadrant of the circle corresponds to different human-exoskeleton interaction modes determined by the variation of the normalized total muscle activation ($_{c_1}^{c_2}\Delta \mu$) with respect to the normalized total interaction torque ($_{c_1}^{c_2}\Delta \tau$) between controllers $c_1$ and $c_2$, respectively. The first quadrant (red) indicates increased disagreement between the user and exoskeleton, resulting in an increase in both muscle effort and the total interaction torque. The second quadrant (green) determines the co-adaptation of the user toward participating in the motion as much as possible and leading the motion. The third quadrant (blue) denotes the decrease in total interaction torque and the total muscle effort, associated with the decrease in human-exoskeleton disagreement. Finally, the fourth quadrant (orange) denotes the condition at which the user yields control of the motion to the exoskeleton and minimally activates their muscles. In this case, muscle activation decreases while the total interaction increases since the exoskeleton has to carry the user's body (passive dynamics) in addition to the exoskeleton dynamics.}
	\label{fig1}
\end{figure*}

To determine the appropriate control strategy for human augmentation and rehabilitation applications, it is crucial to understand human-exoskeleton adaptation as an indicator of how individuals respond to specific exoskeleton control strategies concerning shared motion control~\cite{Jackson2019}. In power augmentation, the ideal scenario involves users adapting to a strategy in which they contribute primarily by guiding the motion, without physical exertion~\cite{Durandau2022}. The exoskeleton takes the responsibility of moving the human body by applying interaction torques or forces to the human body, as demonstrated by reduced muscle activity or metabolic rates~\cite{Franks2021}. Conversely, in rehabilitation, users must often be guided to increase their muscle activity and actively engage in motion control~\cite{Banala2007}. The human-exoskeleton interaction torques exhibit a dual behavior in this context. When the user performs the motion correctly, the exoskeleton must transparently follow the user, resulting in zero interaction torques~\cite{Kuecuektabak2024}. However, when motion correction is required, the exoskeleton controller should intervene. This intervention creates a conflict necessitating an increase in the interaction torque to rectify the motion. Neither muscular effort nor interaction torques alone can discern the aforementioned conditions. For example, an increase in muscular effort may stem from human-exoskeleton disagreement~\cite{Shushtari2021}, while it can also signify that the human user is engaged in walking and relies on their motor capacity rather than on exoskeleton assistance. Therefore, to compare different controllers in such cases, interaction torque needs to be considered alongside muscular effort. A low level of interaction torque coupled with higher muscular effort suggests no physical disagreement, indicating that the exoskeleton is following the user and the user is walking with minimal assistance. Otherwise, a higher level of interaction torque along with high muscular effort indicates that the user and exoskeleton do not share the same desired motion patterns, and they are fighting for control~\cite{Losey2018}. Therefore, determining the suitability of a controller for either power augmentation or rehabilitation applications necessitates co-analysis of muscular effort and interaction torque.

Inspired by the above reasoning, we propose evaluating human-exoskeleton physical interaction by co-analyzing the variation of muscular effort ($\Delta \mu$) and interaction torques ($\Delta \tau$) as a 2D random variable in the $\Delta \tau-\Delta \mu$ space, which draws the Interaction Portrait (IP), i.e., the distribution of the $\Delta \tau-\Delta \mu$ random variable. According to \autoref{fig1}, we show that the phase of the IP distribution (in polar coordinates) indicates whether the human-exoskeleton interaction is developing toward yielding motion control to the human, to the exoskeleton, or toward an increase in human-exoskeleton physical disagreement. Moreover, the temporal analysis of the IP phase reveals how this interaction evolves over time. 
This metric is utilized to compare a recently proposed feedforward controller, developed based on required joint torque estimation using kinematic states called Hybrid Torque Controller (HTC)~\cite{Dinovitzer2023}, with a novel feedforward controller proposed in this paper called Adaptive Model-Based Torque Controller (AMTC), that learns the user's desired trajectory and employs the exoskeleton's dynamical model to generate feedforward torques. The variation of muscle activity and interaction torques in these controllers are compared to a baseline time-based torque controller (TBC), using the IP. This analysis aims to identify how humans adapt to each controller and examine the implications of each controller for human power augmentation and rehabilitation applications.
IP analysis of the data collected from nine able-bodied participants who walked on a treadmill with an exoskeleton controlled by TBC, HTC, and AMTC reveals that the HTC controller led participants to rely more on the exoskeleton, ideal for power augmentation. On the other hand, the AMTC controller increased user engagement by making the exoskeleton less intrusive, advantageous for rehabilitation and gait training. The Results section starts with a formal analysis of interaction torque, EMG and metabolic rate. Then, we present the IP analysis and discuss its results in the next section.

\section{Results}\label{sec2}
\subsection{Initial Processed Data for a Representative Participant}

\begin{figure*}[h!]
	\centering	
	\scalebox{1}{\includegraphics[trim=50 0 0 0]{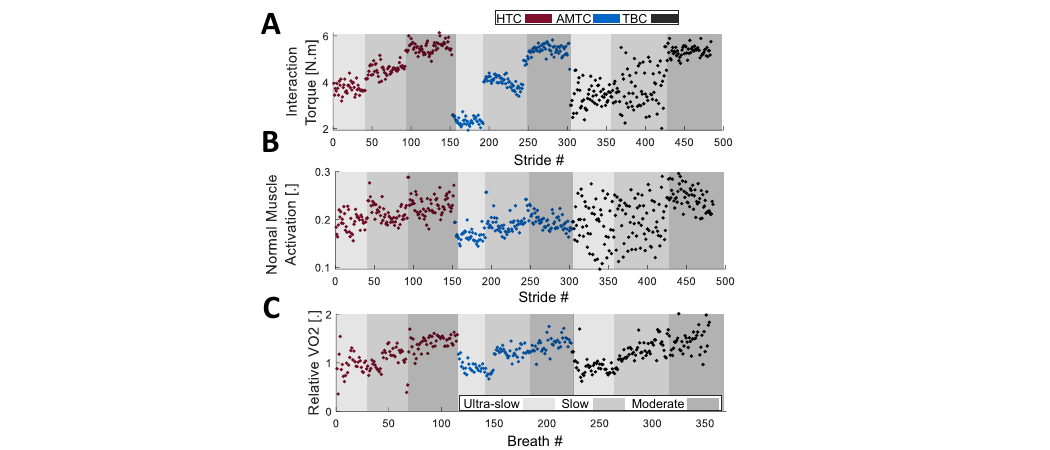}}
	\caption{{A portion of a typical participant's experimental data}; for ease of visualization and interpretation, the interaction torque at the right hip and activation of one of the muscles are illustrated along with the relative oxygen uptake. \textbf{(A)} The mean absolute interaction torque at the right hip at each stride with each controller and speed for Participant \#1. \textbf{(B)} Normal muscle activation for the Gastrocnemius Medialis at the right leg. \textbf{(C)} Relative oxygen uptake for each breath for each controller and speed. The oxygen uptake has increased with the increase in treadmill speed.}
	\label{fig_2}
\end{figure*}

\autoref{fig_2} showcases some of the signals obtained from a typical participant (Participant \#1) during robot-assisted treadmill walking with different controllers. \myref{fig_2}{A} shows the mean absolute interaction torque at the right hip joint (as one of the four joints we computed interaction torque about) for each stride. The interaction torque increases with an increase in treadmill speed across all blocks. During ultra-slow and slow walking, AMTC interaction torques are smaller than those of HTC and TBC. \myref{fig_2}{B} shows Gastrocnemius Medialis (GM) activation ($\mu_{c,v,m}$), selected for visualization here as an example from seven muscles recorded on each leg, for each stride. Similar to the interaction torques, muscular effort increases at higher treadmill speeds, mostly during the HTC and TBC blocks. In the AMTC block, however, GM's muscular effort does not change from slow to moderate speeds. Moreover, the muscular effort is smaller during the AMTC block compared to the two other blocks. The relative oxygen uptake ($\eta_{c,v}$) during the experiment for breath cycles is presented in \myref{fig_2}{C}. As anticipated, the oxygen uptake increases with an increase in treadmill speed in the HTC, AMTC, and TBC blocks.

\subsection{Overall Performance Analysis}

\begin{figure*}[h!]
	\centering	
	\scalebox{.4}{\includegraphics[trim=40 0 0 0]{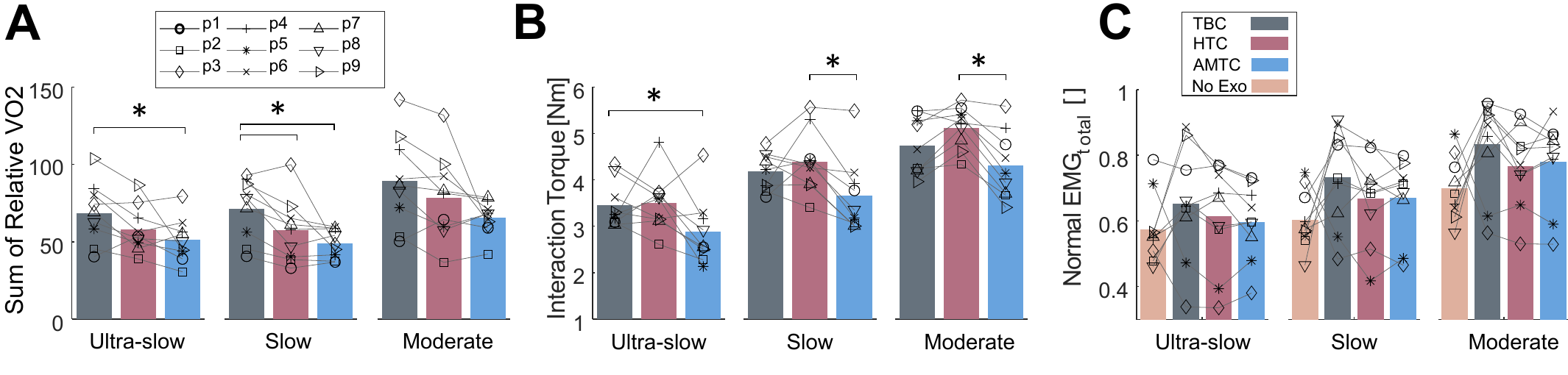}}
	\caption{{The average performance metrics for each treadmill speed and controller across participants}. \textbf{(A)} The sum of the relative oxygen uptake across all the strides for each speed in each controller block graphed for each participant. The bars show the average of the sum of the oxygen uptake across all participants. Similarly, the average total absolute value of the human-exoskeleton interaction and total normalized muscle effort are graphed in \textbf{(B)} and \textbf{(C)}, respectively.}
	\label{fig_3}
\end{figure*}

\myref{fig_3}{A} shows the sum of the oxygen uptake for participants for each of the TBC, HTC, and AMTC blocks during ultra-slow, slow, and moderate-speed walking. TBC and AMTC have the highest and lowest metabolic rate at all walking speeds, respectively. The AMTC-resultant metabolic rate is significantly less than other controllers, at ultra-slow and slow walking, where AMTC resulted in 22.9\%±17.1 (Friedman: $p<$0.03, Wilcoxon signed rank: $p_{_{TBC,AMTC}}<$0.01) and 28.7\%±12.7 (Friedman: $p<$0.005, Wilcoxon signed rank: $p_{_{TBC,AMTC}}<$0.003) decrease in the total oxygen uptake, respectively. 
The total mean absolute interaction torque is similarly illustrated for the participants in \myref{fig_3}{B}. AMTC has the lowest interaction torque compared to TBC and HTC, indicating the least disagreement between the exoskeleton assistance and the user desired motion. With respect to the TBC, AMTC shows 17.1±12.5\%, 12±15\%, and 9.2±7.7\% of reduction in human-exoskeleton total interaction in ultra-slow, slow, and moderate-speed walking, respectively. The difference is statistically significant at the ultra-slow walking (Friedman: $p<$0.04, Wilcoxon signed rank: $p_{_{TBC,AMTC}}<$0.01). Compared to HTC, AMTC shows 19.8±21.1\%, 17.9±10.1\%, and 18.1±9.9\% reduction in human-exoskeleton total interaction. These differences are statistically significant in the case of slow (Fridman: $p<$0.03, Wilcoxon signed rank: $p_{_{HTC,AMTC}}<$0.004) and moderate-speed (Fridman: $p<$0.0008, Wilcoxon signed rank: $p_{_{HTC,AMTC}}<$0.004) walking.  
\myref{fig_3}{C} shows the total muscle effort for participants' right legs during ultra-slow, slow, and moderate-speed walking across the three different controllers. Natural walking without the exoskeleton has the lowest total muscle effort compared to other cases in which the exoskeleton is involved. This is expected as wearing the exoskeleton adds about 17 kg of extra weight to the body resulting in higher muscle effort. Among the three controllers, TBC has the highest total muscular effort at all speeds. AMTC and HTC's total muscular effort are close in all cases while AMTC is slightly lower and higher in ultra-slow and moderate speeds, respectively. None of the identified differences are statistically significant.  

\subsection{Interaction Portrait Analysis}

\begin{figure*}[h!]
	\centering	
	\scalebox{.4}{\includegraphics[trim=40 0 0 0]{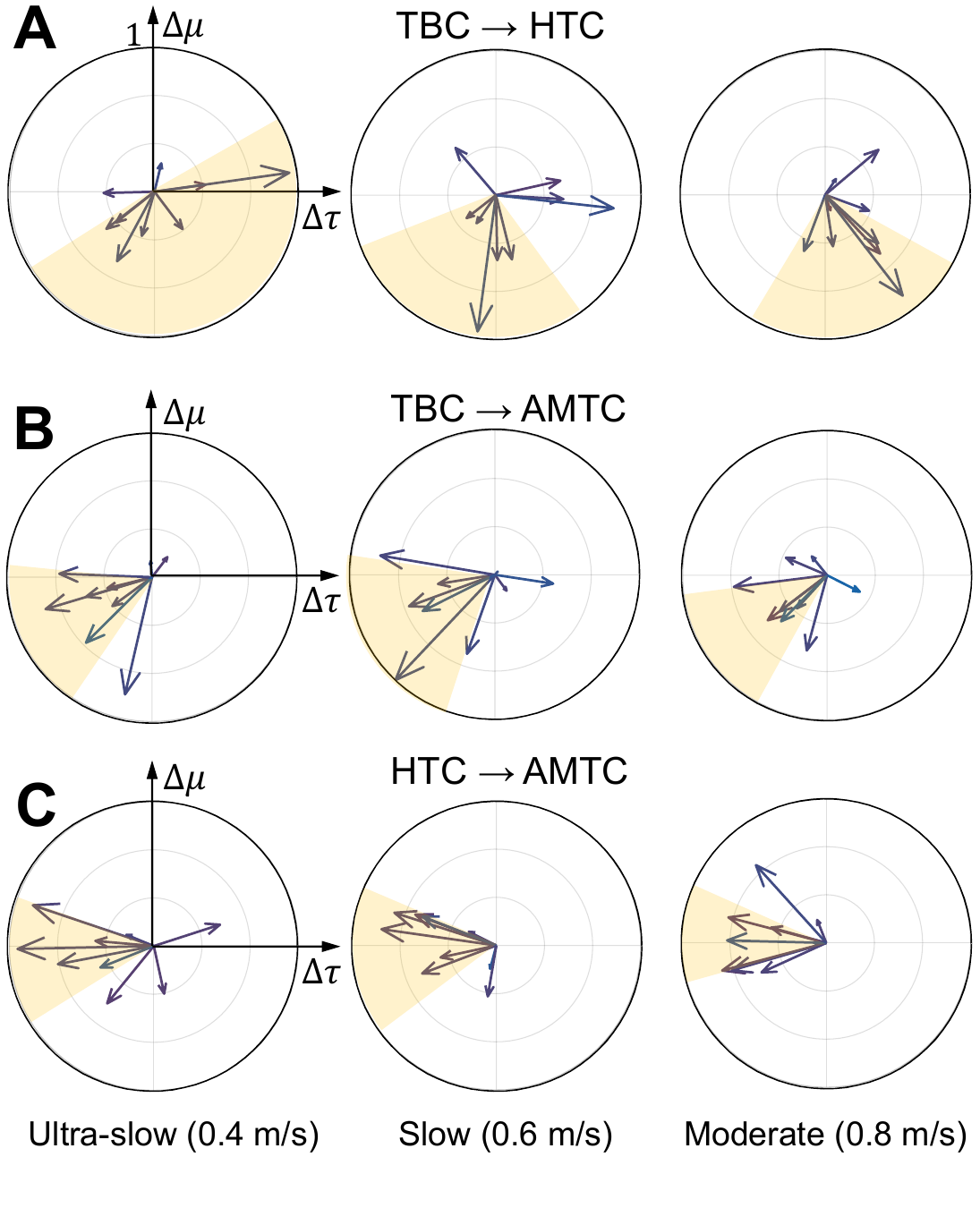}}
	\caption{{Comparing the Average Interaction Portrait for each Pair of Controllers}. The average Interaction Portrait (IP) depicted according to the average total muscle effort and the average total human-exoskeleton interaction for each participant computed at each of the ultra-slow, slow, and moderate-speed walking for the TBC$\rightarrow$HTC, TBC$\rightarrow$AMTC, and HTC$\rightarrow$AMTC illustrated in \textbf{(A)}, \textbf{(B)}, and \textbf{(C)}, respectively. The yellow areas denote the area between the 25 and 75 percentiles.}
	\label{fig_4}
\end{figure*}

\autoref{fig_4} compares the examined controllers one by one in ultra-slow, slow, and moderate-speed walking by illustrating the average change in the max-normalized total muscular effort with respect to the change in the max-normalized total interaction torque (the average IP is plotted by a single vector for each participant). To compare the IP results of participants more easily,\myref{fig_4}{A} shows the average IP for HTC compared to TBC, denoted as $[_{TBC}^{HTC}\Delta \tau_v ^{tot} , _{TBC}^{HTC} \Delta \mu_v^{tot}]$. In ultra-slow and slow walking, participants adapt differently to HTC compared to TBC, as the average IP, indicated by vectors, are spread in all quadrants. In moderate speed, however, the majority of average IP vectors fall within the fourth quadrant, indicating that most participants yield control to the HTC-controlled exoskeleton. This observation aligns with participants exhibiting the lowest total muscular effort when walking with an HTC-controlled exoskeleton, as shown in \myref{fig_3}{C}. Similarly, \myref{fig_4}{B} presents the average IP vectors of AMTC compared to TBC, denoted by $[_{TBC}^{AMTC}\Delta \tau_v ^{tot} , _{TBC}^{AMTC} \Delta \mu_v^{tot}]$. Unlike HTC, AMTC effectively decreases user-exoskeleton disagreement at all tested speeds, as indicated by the majority of average IP vectors concentrated in the third quadrant. To further examine, \myref{fig_4}{C} compares the HTC and AMTC controllers, denoted as $[_{HTC}^{AMTC}\Delta \tau_v ^{tot} , _{HTC}^{AMTC}\Delta \mu_v^{tot}]$. It shows that at all walking speeds, participants lean more toward contributing to the motion rather than yielding control to the exoskeleton, as most of the average IP vectors fall around the border of the second and third quadrants.

\subsection{Individual Adaptation Strategy}

\begin{figure*}[h!]
	\centering	
	\scalebox{.7}{\includegraphics[trim=10 0 0 0]{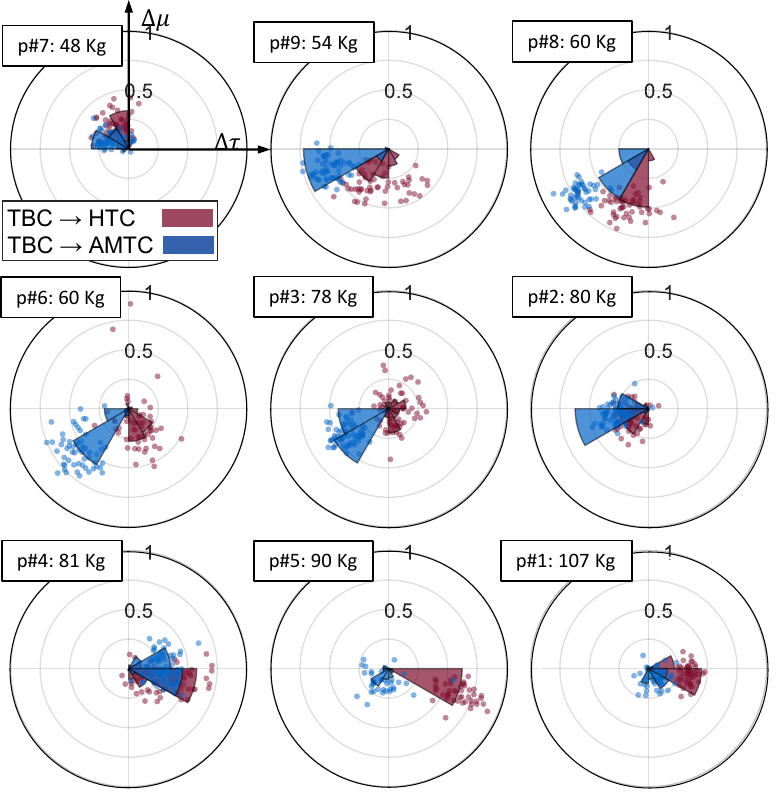}}
	\caption{{Comparison of the Interaction Portrait distribution between TBC$\rightarrow$HTC and TBC$\rightarrow$MTBC}. Interaction portrait distribution for HTC and AMTC blocks with respect to the average total muscle effort and total interaction torque across all strides during the TBC block graphed for each participant plotted for moderate speed walking. The radius of data points is normalized with respect to the maximum radius computed across all participants’ strides. Participants are arranged increasingly according to their body mass. The polar histograms show the concentration intensity of the depicted points. Each bin of the histogram covers $\pi/6$ rad.}
	\label{fig_5}
\end{figure*}

We analyzed the dependency of user adaptation strategy formed in interaction with each of the HTC and AMTC cases with respect to the TBC. To this end, we computed the average total muscular effort and interaction torque in TBC. We then computed the total muscular effort and interaction torque for each stride in the HTC and AMTC cases. For each stride, the difference in total muscle effort and interaction torque is then computed with respect to the TBC. \autoref{fig_5} shows the stride-wise IP distribution of the TBC$\rightarrow$HTC ($[_{TBC}^{HTC}\Delta \tau_{v,s} , _{TBC}^{HTC}\Delta \mu_{v,s}]$ in red), and TBC$\rightarrow$AMTC ($[_{TBC}^{AMTC}\Delta \tau_{v,s} , _{TBC}^{AMTC}\Delta \mu_{v,s}]$ in blue) for each participant separately during moderate speed walking. The graphs are ordered from left to right and top to bottom, corresponding to a monotonic increase in participants' body mass. According to the graph, except for participant \#7, our lighter participants leaned towards contributing more to the gait and leading the motion, with either HTC or AMTC, as their IP is consistently distributed in the third quadrant. Conversely, heavier participants in our experiment showed a tendency to relinquish control and passively follow the exoskeleton. This tendency is more pronounced in the case of the two heaviest participants who consistently adopted such a strategy in the case of HTC, with their IP distribution falling in the fourth quadrant, while their IP distribution for the AMTC is spread over a wider area. The strength of the adopted co-adaptation strategy is proportional to the radius of the IP distribution. Accordingly, the higher radius of TBC$\rightarrow$AMTC IP distribution compared to TBC$\rightarrow$HTC in the case of participants \#9, 8, 6, and 3 reveals that AMTC-controlled exoskeleton led users to adopt a more consistent strategy compared to the HTC case. Participants \#5 and 1, as exceptions, adopted a more consistent strategy during HTC. \autoref{fig_5} also reveals that Participant \#1 did not adopt a strong strategy in either the HTC or AMTC cases.

\section{Discussion}\label{sec3}
\subsection{Human Adaptation}

\begin{figure*}[h!]
	\centering	
	\scalebox{1.1}{\includegraphics[trim=20 0 0 0]{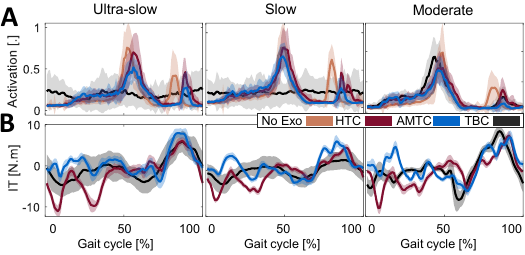}}
	\caption{Muscle activation and interaction torque profile with respect to the gait phase. \textbf{(A)} The average normalized muscle activation pattern for the TBC, HTC, and MBTC blocks for the right Gastrocnemius Medialis during ultra-slow, slow, and moderate-speed walking. The shaded area represents the standard deviation of the muscle activation about their mean value. Similarly, the average human-exoskeleton interaction torque at the right hip is plotted in \textbf{(B)}.}
	\label{fig_6}
\end{figure*}

An interesting observation in \autoref{fig_2} is the variation in metrics associated with the TBC controller. For instance, Participant \#1 experienced varying interaction torques at ultra-slow and slow walking speeds, as shown in \myref{fig_2}{A}. Conversely, at moderate-speed walking, the variation in the interaction torque decreases, demonstrating a more consistent performance by the participant. A similar observation is evident in the case of GM muscle activation (\myref{fig_2}{B}). For further investigation, \autoref{fig_6} illustrates the average profile of GM activation and hip interaction torque, temporally normalized with respect to the gait phase at ultra-slow, slow, and moderate-speed walking. Both metrics exhibit a large standard deviation (shaded area) during ultra-slow and slow walking, while the standard deviation significantly drops during moderate-speed walking. This indicates an inconsistency in the participant's performance in both spatial and temporal aspects.

The inherent difference between the TBC and the other two controllers can explain this observation. During the TBC block, the feedforward torque is generated regardless of the user's performance, and the exoskeleton has no capacity for adapting to the user. Therefore, it is entirely up to the user to adapt to the torque delivered by the exoskeleton. In ultra-slow and slow walking, the participant was unable to adjust their gait timing to the exoskeleton, resulting in inconsistent interaction with the exoskeleton. At the higher speed, however, the user was able to synchronize their gait with the exoskeleton assistance and, therefore, adopted a solid interaction strategy, which emerges as a lower variation in all the above metrics. This has not been the case for all participants, as six of them were not successful in effectively synchronizing their walking to the TBC-controlled exoskeleton assistance at any speed. These results highlight the importance of human-robot co-adaptation, which is not achievable with a time-based controller.

\subsection{Importance of IP Analysis}

\begin{figure*}[h!]
	\centering	
	\scalebox{1}{\includegraphics[trim=55 0 0 0]{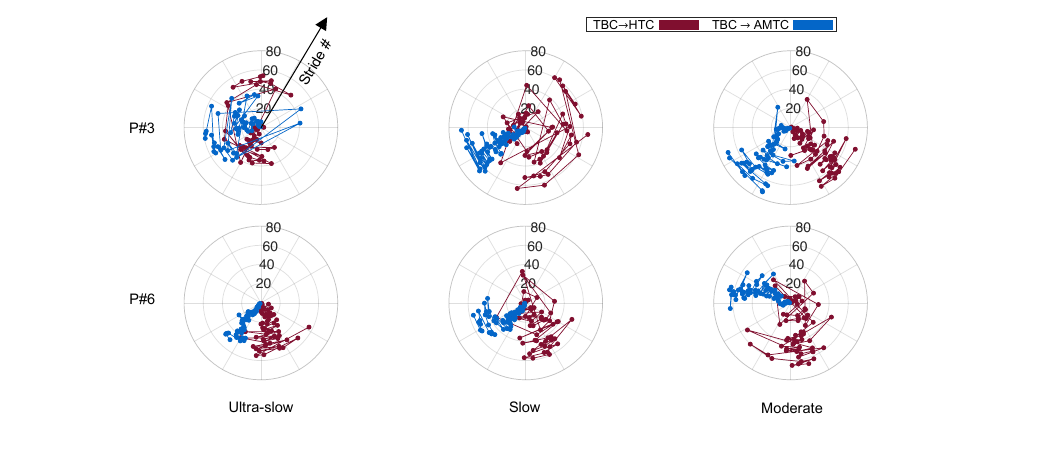}}
	\caption{Evolution of IP phase at each stride at different walking speeds for two sample participants. The top and bottom rows depict IP phase evolution at ultra-slow, slow, and moderate speed walking for each of the TBC$\rightarrow$HTC and TBC$\rightarrow$AMTC cases for participants \#3 and \#6, respectively.}
	\label{fig_7}
\end{figure*}

In \myref{fig_3}{A}, the lower total muscular effort during walking with the HTC and AMTC compared to TBC is consistent with their lower metabolic rate with respect to the TBC, indicating that both controllers reduce the walking effort more than the TBC-controlled exoskeleton. However, a similar trend has not been observed in the total interaction torque, where HTC resulted in higher interaction torque compared to the TBC. This indicates that even though HTC and AMTC both reduced users’ metabolic rate and muscular effort, they encouraged users to adopt two different interaction strategies. IP analysis further investigates the effect of each controller on the performance of each participant by analyzing the difference in total muscular effort with respect to the difference in total human-exoskeleton interaction torque. \autoref{fig_4} reveals that the decrease in metabolic rate occurred based on two different human adaptation strategies. In the HTC controller, users relinquished control to the exoskeleton, passively following the exoskeleton's motion. In contrast, AMTC encouraged users to lead the motion more actively. Our IP analysis suggests that the HTC controller is particularly well-suited for applications requiring power augmentation, such as in industrial settings for workers or healthcare environments for nurses. In these contexts, the primary goal is to minimize human exertion, thereby enhancing operational capabilities and safety. Conversely, the AMTC controller shows greater promise in rehabilitation contexts for individuals with residual motor functions, such as those with incomplete spinal cord injuries or post-stroke conditions. Here, the imperative is to actively involve the user in task execution, thereby amplifying their motor functions and accelerating recovery processes.

\autoref{fig_4} also demonstrates that different participants adopted a more consistent strategy with AMTC compared to the HTC, as the average IP vectors across participants exhibit lower variation with AMTC compared to the HTC controller.

IP distribution itself can also shed light on the strength of the adopted strategy in each participant depending on the radius of the vectors forming the IP. This is more evident in \autoref{fig_5} where each point of IP represents the difference in muscle effort and interaction torque obtained for each stride. As an example, the IP analysis in \autoref{fig_5} reveals that Participant \#3 (2nd row, 2nd column) and Participant \#6 (2nd row, first column) adopted the same interaction strategy with the AMTC controller compared to the TBC. This is, nevertheless, more significant for Participant \#6 due to the larger radius of the distributed points compared to those of Participant \#3.

Using the IP analysis, it is also possible to track the evolution of the adopted strategy across the stride at each walking speed. \autoref{fig_7}, as an example, shows the evolution of the IP phase for each of the TBC$\rightarrow$HTC and TBC$\rightarrow$AMTC comparisons for participants \#3 and \#6, respectively. Implied by the large variation in the IP phase, Participant \#3 did not converge to a consistent interaction strategy with neither the HTC nor AMTC controllers during ultra-slow walking. This lack of convergence to a consistent interaction with the exoskeleton agrees with our difficulty in maintaining smoothness and continuity of ultra-slow movements, possibly explained by limitations of dynamic primitives that often lead to segmented movements (Park, et al., 2017). In slow walking, however, the user adopted a more consistent strategy using AMTC, evidenced by low variations in the IP phase. In the case of the HTC controller, the participant’s strategy remains inconsistent. Only in the case of moderate-speed walking was Participant \#3 able to converge to consistent interaction strategies both with the AMTC and the HTC controllers. Our IP analysis in this case shows that AMTC decreased the human-exoskeleton interaction, but the user did not completely obtain the motion control or yield the motion to the exoskeleton, as the IP phase is still in the third quadrant. In thecase of the HTC controller, the user has relied more on the exoskeleton assistance since the IP phase is primarily concentrated in the 4th quadrant. Participant \#6, in contrast to Participant \#3, converged to a consistent interaction with the exoskeleton at all three walking speeds. The HTC controller, regardless of the walking speed, has guided the participant to rely more on the exoskeleton as the IP phase is mostly concentrated on the border of the 3rd and 4th quadrants. In the case of the AMTC controller, however, we observe that as the gait speed increases, the user strategy develops more toward leading the gait and contributing to motion control, evidenced by an 83-degree shift in the average IP phase in moderate-speed walking compared to ultra-slow walking.

These results showcased the ability of IP analysis to provide an objective comparison of different exoskeleton controllers, the adopted interaction strategy by the user, as well as evaluating user-exoskeleton co-adaptation. Besides offline analysis, IP provides designers with a quantitative metric that can be tuned in a human-in-the-loop optimization setting to tailor the exoskeleton controller to the unique requirements of each application or participant.

\section{Conclusion}\label{sec5}
We proposed a new metric for the analysis of human-exoskeleton interaction (Interaction Portrait) and employed it in investigating the effect of three feedforward controllers in enhancing human-exoskeleton interaction during assisted treadmill walking at different speeds. Through interaction portrait analysis, we found that the HTC controller demonstrated a more suitable performance for power augmentation along with reducing muscle activation and metabolic cost. On the other hand, the AMTC controller, also proposed in this study, proved to be more suitable for rehabilitation applications, as it promoted user reliance on their own muscular capacity by making the exoskeleton transparent. 

Furthermore, we observed that the human adaptation pattern facing each of the HTC and AMTC controllers was influenced by the participants' weight. Individuals with lower weight tended to take control with the AMTC controller, while heavier participants were more inclined to relinquish control when interacting with the HTC-controlled exoskeleton. IP analysis has been only performed to evaluate the able-bodied individuals’ interaction with exoskeletons. As the next step of this research, we plan to analyze the interaction portrait of motor-impaired individuals with exoskeletons equipped with different assist as needed controllers to provide a meaningful and objective comparison of these controllers.    

\section{Methods}\label{sec4}
\subsection{Feedforward Control Strategies}
This section describes three controllers tested in our study. 
\begin{figure*}[h!]
	\centering	
	\scalebox{.7}{\includegraphics[trim=0 0 0 0]{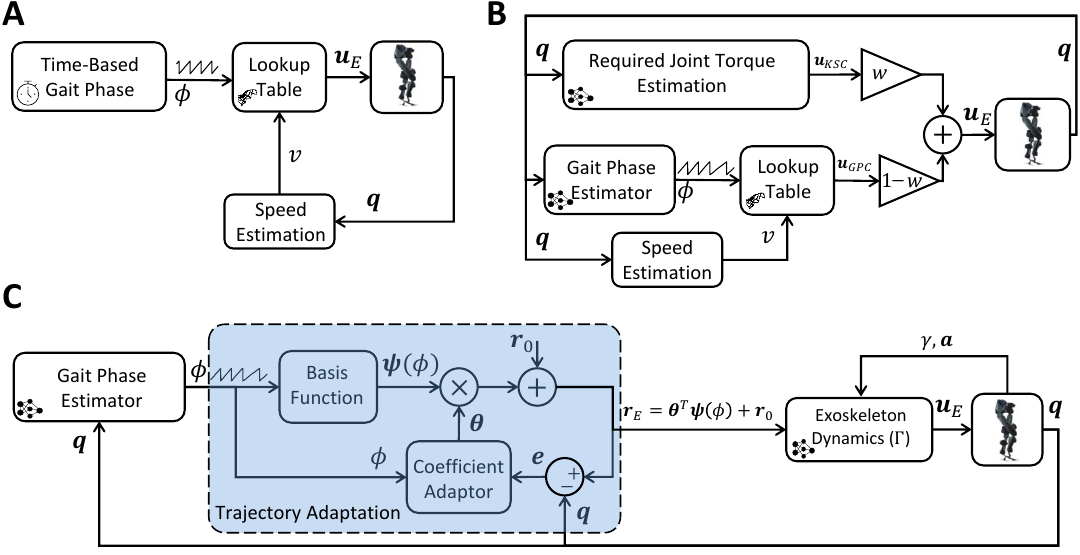}}
	\caption{Block diagram of TBC, HTC, and AMTC controllers. \textbf{(A)} Block diagram of the Time-Based Controller (TBC). A time-based gait phase along with the estimated gait speed are fed into a lookup table to determine the applied torque to the exoskeleton joints according to joint torque data recorded from the exoskeleton during high-gain joint control with the user passively following the exoskeleton (with the minimum voluntary contribution to the gait). \textbf{(B)} The diagram of the Hybrid Torque Controller (HTC) consisted of a data-driven required joint torque estimator along with a look-up table-based torque controller similar to the TBC. In this case, however, the gait phase is determined according to the exoskeleton states rather than time. The torque from the two different pipelines is finally combined with the weight of $w$ and $1-w$ to form the applied torque to the exoskeleton. \textbf{(C)} Block diagram of the Model-Based Torque Controller (AMTC). The gait phase is estimated according to the exoskeleton joint angles and then fed into a trajectory adaptation block which learns the joint trajectory of the participant in real time and uses that trajectory as the reference for the exoskeleton to be fed into the forward dynamics of the exoskeleton to determine the feedforward joint torques.}
	\label{fig8}
\end{figure*}

\subsubsection{Time-Based Torque Controller}

\myref{fig8}{A} depicts the block diagram of the Time-Based Torque Controller (TBC). This controller utilizes lookup tables to determine the desired joint torques $\bm{u}_E$ by considering both the gait phase ($\phi(t)$ and the estimated gait speed ($v$). To construct the lookup table, measurements of exoskeleton joint torques were taken at different speeds while a participant walked with the exoskeleton governed by a high-gain PD controller. The participant was asked to exert the minimum voluntary effort during walking. The joint trajectories were controlled based on reference trajectories derived from the participant's walking without the exoskeleton. The input gait phase to the lookup table was obtained by dividing the stride length, calculated using the exoskeleton joint angles ($\bm{q}$), by the stride time updated at each heel strike. For this controller, the gait phase ($\phi(t)$) was generated based on the desired gait speed, following the formula $\phi(t) = mod(t,T)$, where $T$ represents the average measured stride time when the participant walked without the exoskeleton at the desired gait speed. For more detailed information about the construction of the lookup table and the gait speed estimator, refer to \cite{Dinovitzer2023}.

\subsubsection{Hybrid Torque Controller}

\myref{fig8}{B} presents the block diagram of the Hybrid Torque Controller (HTC), which combines the torque outputs of two distinct controllers \cite{Dinovitzer2023}: the Kinematic State Dependent Controller (KSC) and the Gait Phase Dependent Controller (GPC). The KSC incorporates an Artificial Neural Network (ANN) that calculates the required biological torque for the user based on the kinematic measurements of the exoskeleton \cite{Dinovitzer2023a}. On the other hand, the GPC adopts the same structure as the TBC mentioned earlier, however, instead of relying on a time-based gait phase, it utilizes a real-time estimation of the gait phase derived from the kinematic measurements \cite{Shushtari2022}. The outputs of these two controllers are linearly combined in the HTC: $\bm{u}_E= w \bm{u}_{KSC}+ (1-w) \bm{u}_{GPC}$, where $w$ represents the weight assigned to the KSC output. 

\subsubsection{Adaptive Model-based Torque Control}

The Adaptive Model-based Torque Control (AMTC) depicted in \myref{fig8}{C}, leverages the estimated dynamics of the exoskeleton to generate joint torques for control. According to \cite{Shushtari2023}, the interaction between the Indego exoskeleton and the human in the sagittal plane can be described by the following dynamical model: 
\begin{equation} \label{eq_1}
	\bm{\Gamma}(\gamma,\dot{\gamma},\ddot{\gamma},\bm{{a}},\bm{q},\bm{\dot{q}},\bm{\ddot{q}}) = \bm{u}_E + \bm{u}_{int},
\end{equation}

\noindent where $\gamma \in \mathbb{R}$ and $\bm{a} \in \mathbb{R}^2$ denote the exoskeleton thigh segment angle with respect to the gravity vector and its acceleration in the sagittal plane, respectively, $\bm{q} = [q_{h,r}; q_{k,r}; q_{h,l}; q_{h,l}]\in\mathbb{R}^4$ represent the exoskeleton hip and knee joint angles, $\bm{u}_E \in \mathbb{R}^4$ denote the exoskeleton applied motor torques, and $\bm{u}_{int} \in \mathbb{R}^4$  represents the torques arising from the human-exoskeleton interaction. 
The AMTC controller employs a dynamic compensatory approach. In a dynamic compensator, the exoskeleton's applied torques are set equal to the exoskeleton's passive dynamics ($\bm{u}_E = \bm{\Gamma}(\gamma, \dot{\gamma}, \ddot{\gamma}, \bm{{a}},\bm{q}, \bm{\dot{q}}, \bm{\ddot{q}})$) which ideally results in the transparency of the exoskeleton ($\bm{u}_{int} = 0$). However, in AMTC, instead of using the current joint angles, the desired joint angles are used to compute the exoskeleton torques: 
\begin{equation}
	\bm{u}_E = \bm{\Gamma}(\gamma,\dot{\gamma},\ddot{\gamma},\bm{{a}},\bm{r}_E,\bm{\dot{r}}_E,\bm{\ddot{r}}_E),
\end{equation} 
\noindent where $\bm{r}_E \in \mathbb{R}^4$ are the desired exoskeleton joint angles. This approach allows the exoskeleton to be transparent only when the user's joint motions align with the desired trajectories. Otherwise, the exoskeleton will assist or resist the user's motion depending on the consistency between the user's desired motion and the exoskeleton's desired trajectories.
To ensure that the exoskeleton always assists the user and avoids resisting their motion, the reference trajectory of the exoskeleton needs to be synchronized temporally with the user's desired motion and matched spatially to their intended gait pattern. Temporal synchronization is achieved by defining the exoskeleton's reference trajectory as a function of the estimated gait phase computed from the exoskeleton's joint angles ($\bm{r}_E(t) = \bm{r}_E(\phi(\bm{q}))$). Spatial consistency is ensured by adapting the reference trajectory based on the minimization of the error between the exoskeleton's reference trajectory and its current joint angles ($\bm{e} = \bm{r}_E - \bm{q}$). At each joint, the reference trajectory is defined as ${r}_E = {r}_0(\phi) + \Delta(\phi)$, where ${r}_0$ represents the initial reference trajectory of the exoskeleton, and $\Delta(\phi) = \bm{\theta}^T \times \bm{\psi}(\phi)$ is the modification term. Here, $\bm{\psi(\cdot)}$ represents the Fourier series basis functions with up to $m$ harmonics, and $\bm{\theta}$ represents the coefficients of the Fourier series that are adapted similar to \cite{Shushtari2021}. The adaptation rule is then derived to minimize $J = 0.5e^2$ using Gradient Descent (with a learning rate of $\epsilon$): 
\begin{equation}
	\dot{\bm{\theta}} =-\epsilon \frac{\partial{J}}{\partial{\bm{\theta}}}=-\epsilon \frac{\partial{J}}{\partial{e}} \times \frac{\partial{e}}{\partial {r_E}} \times \frac{\partial{r_E}}{\partial{\Delta}} \times \frac{\partial{\Delta}}{\partial{\bm{\theta}}} =-\epsilon e \bm{\psi}(\bm{\phi}).
\end{equation}

\subsection{Experimental Setup}

\begin{figure*}[h!]
	\centering	
	\scalebox{1}{\includegraphics[trim=0 0 0 0]{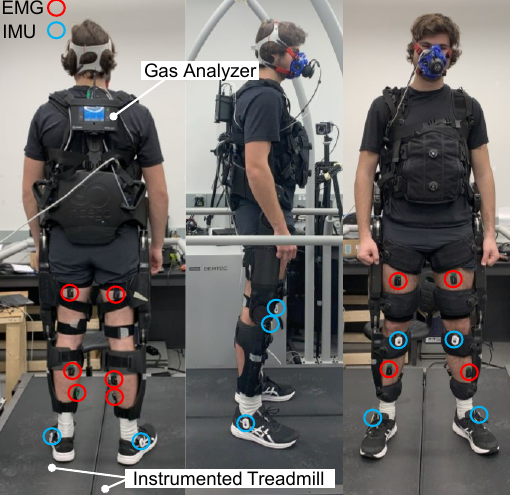}}
	\caption{Dorsal, lateral, and frontal view of a participant with the Indego exoskeleton with active hip and knee joints. The participant is standing on the Bertec treadmill with two speed-controlled belts equipped with individual loadcells underneath each of them for GRF monitoring. Muscle activation is measured from both right and left leg muscles using EMG sensors. Gait up IMUs are clipped to the outer side of the shoes right bellow ankle joints to measure the spatiotemporal parameters of gait. Oxygen uptake of the participant is measured and recorder at each breath through a mask connected to the gas analyzer device carried at the back of the participant.}
	\label{fig9}
\end{figure*}

The experimental setup comprised a split-belt instrumented treadmill (Bertec, US), with force plates under each belt, providing ground reaction forces (GRF) under each belt, a lower limb exoskeleton (Indego, Parker Hannifin, US) with actuated hip and knee joints, fourteen wireless EMG sensors (Trigno, Delsys, USA), and a COSMED K5 wearable metabolic system (Albano Laziale, Roma, Italy) to measure oxygen uptake (VO2). Additionally, four Physilog IMUs (Physilog 6s, Gait Up SA, CH) and a COSMED K5 wearable metabolic system (Albano Laziale, Roma, Italy) are employed to measure gait spatiotemporal parameters and oxygen uptake (VO2), respectively. The exoskeleton, IMUs, load cells, and EMG sensors have sampling rates of 200 Hz, 128 Hz, 1000 Hz, and 2000 Hz, respectively. Following appropriate skin treatment, EMG sensors were placed on Gluteus Maximus, Biceps Femoris, Rectus Femoris, Vastus Medialis, Gastrocnemius Medialis, Soleus, and Tibialis Anterior muscles of each leg. Spatiotemporal gait parameters such as Minimum Toe Clearance (minTC), Maximum Heel Clearance (maxHC), Stance Time Percentage, and Stride Length are computed using the Physilog sensors, as validated in previous studies \cite{carroll2022validation, schwameder2015validation}. Before recording data for each subject, the COSMED K5 underwent a calibration process following a standardized procedure. \autoref{fig9} illustrates the exoskeleton and the placement of the sensors within the experimental setup.

\subsection{Experimental Protocol}

\begin{figure*}[h!]
	\centering	
	\scalebox{1}{\includegraphics[trim=0 0 0 0]{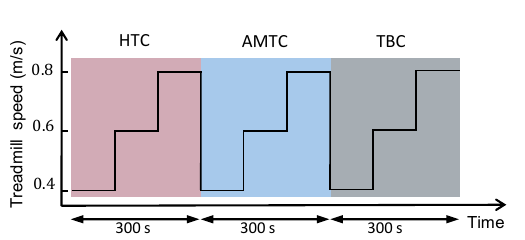}}
	\caption{{Treadmill speed changes while experimenting with the HTC, AMTC, and the TBC controller}. The order of the controllers was specific to participant \#1 and varied for other participants. Participants walked with each controller for 300 seconds divided by three 100-second walking periods during each the treadmill speed was set to ultra-slow (0.4 m/s), slow (0.6 m/s), and moderate (0.8 m/s) speeds.}
	\label{fig10}
\end{figure*}

The experiment consisted of three blocks, each involving participants walking on the treadmill with the exoskeleton controlled by one of three different controllers: TBC, HTC, or AMTC. The order of the blocks was varied among participants to mitigate any potential order effects. Within each block, participants walked at three distinct speeds: ultra-slow (0.4 m/s), slow (0.6 m/s), and moderate (0.8 m/s). Each speed lasted for a duration of 100 seconds. For an illustration of the treadmill speed and the order of the controllers applied to the exoskeleton for participant \#1, refer to \autoref{fig10}.

A total of nine able-bodied participants (5 males and 4 females, age: 23.4±4.2 years, mass: 73.6±20.2 kg, height: 176.7±9.6 cm) participated in the study. All participants provided informed written consent prior to the experiments. The study protocol and procedures received ethical approval from the University of Waterloo Clinical Research Ethics Committee (ORE\#41794). The study adhered to the principles outlined in the Declaration of Helsinki.

\subsection{Data Analysis}

\subsubsection{Ground Reaction Force}

Force plates are utilized to measure the ground reaction forces (GRF) in the vertical, lateral, and longitudinal directions. Heel strike events are detected using the vertical GRF, enabling the segmentation of collected data into individual strides. The recorded GRF during no exoskeleton walking is temporally normalized and averaged based on the gait phase as $${\bar{\bm{f}}}_{v}(\phi)=\underset{s\in S_{v}}{\text{mean}}{\bm{f}_{s}(\phi)},$$ where $v\in$\{ultra-slow, slow, moderate\} is the treadmill speed index, $s\in\mathbb{N}$ is the stride index, $S_{v}$ is the set of the strides during no exoskeleton walking with speed $v$, and $\bm{f}_s(\phi)$ is the temporally normalized GRF at stride $s$. To assess the similarity between the ground reaction force profiles with each controller and the natural walking without the exoskeleton, a normalized Pearson correlation is computed as $$\bm{x}_{c,v,s}=\{\text{Corr}{(\bm{f}_{s}(\phi),{\bar{\bm{f}}}_{v}(\phi))}|s\in S_{c,v}\},$$ where $S_{c,v}$ is the set of all strides during walking with exoskeleton at treadmill speed $v$ with controller $c\in\text{\{\ TBC,\ HTC,\ MBTC\}}$. 

\subsubsection{Muscle Activation}

The EMG data was bandpass filtered, with cutoff frequencies of 5 and 500 Hz. The signal was then full-wave rectified and its envelope was computed by applying a moving average with a window of 100 ms. Each EMG signal is normalized by its respective maximum voluntary contraction (MVC), computed for each muscle as the maximum measured contraction during walking on the treadmill (maximum across with and without the exoskeleton walking). The average muscle effort \cite{Pedotti1978,Crowninshield1981} is computed for each muscle at each stride as 
$$\mu_{m,s}=\frac{1}{T_s}\int_{T_s}{e_m^2(t)}dt,$$
where $e(t)$ is the filtered EMG signal, $T_s$ is the duration of stride $s$ computed by heel-strike events obtained from vertical GRF measurements,  $m\in$\{GM, BF, RF, VM, MG, Sol, TA\}, is the muscle index. The muscle effort over all strides with controller $c$ and speed $v$ is, therefore, computed as 
$$\mu_{c,v,m}=\sum_{s\in S_{c,v}}{T_s\mu_{m,s}},$$
where $S_{c,v}$ is the set of all strides taken with controller $c$ and speed $v$. Finally, the total muscular effort ($\mu_{c,v}^{tot}$) was determined as the weighted average of the muscle efforts across all muscles. Physiological cross-sectional areas of muscles, obtained from \cite{Lieber2017}, were employed as the weights for total muscular effort computation. This weighting scheme partly accounts for the differences in force contributions across muscles.

\subsubsection{Interaction Torque}

The human-exoskeleton interaction torque is estimated based on \autoref{eq_1}, given by 
\begin{equation} \label{eq_4}
	\bm{u}_{int} = \bm{\Gamma}(\gamma,\dot{\gamma},\ddot{\gamma},\bm{{a}},\bm{q},\bm{\dot{q}},\bm{\ddot{q}})-\bm{u}_E. 
\end{equation}
The mean absolute interaction torque is then calculated for each joint as: 
$$\tau_{j,s}=\frac{1}{T_s}\int_{T_s} \left|u_{int,j}(t)\right|dt,$$
where $j\in$\{hip$_{\text{right}}$, knee$_{\text{right}}$, hip$_{\text{left}}$, knee$_{\text{left}}$\} is the joint index. The overall absolute interaction torque with each controller is computed as 
$$\tau_{c,v,j}=\ \sum_{s\in S_{c,v}}{T_s\tau_{j,s}}.$$
The total interaction torque ($\tau_{c,v}^{tot}$) is finally obtained for each controller as the average interaction torques across all joints.

\subsubsection{VO2}

VO2 is computed for each breath. To normalize the VO2 measurements across participants, the average VO2 during treadmill walking with no exoskeleton at each speed $v$ is computed as: 
$$	{\bar{\eta}}_{v}=\underset{n\in N{v}}{\text{mean}}\ {\eta}_n,$$
where $n$ is the breath index and $N_v$ is the set of all breaths during walking with no exoskeleton at treadmill speed $v$. The normalized VO2 during exoskeleton walking is then computed as 
$${\hat{\eta}}_{c,v,n}= \{\eta_{n}/{\bar{\eta}}_{v}|n\in N_{c,v}\},$$
where $N_{c,v}$ is the set of all breaths during walking with exoskeleton at treadmill speed $v$ with controller $c$. The sum of the VO2 measurements is also obtained for all of the exhales for each controller as 
$${\hat{\eta}}_{c,v}^{tot}=\sum_{n\in N_{c,v}}{\hat{\eta}}_{c,v,n}.$$ 

\subsubsection{Human-Exoskeleton Interaction Portrait (IP) Analysis}

To examine the impact of each controller on the human-exoskeleton interaction dynamics, we analyzed the changes in total interaction torques ($\Delta \tau$) relative to variations in total muscular effort ($\Delta \mu$) when switching from controller $c_1$ to controller $c_2$ ($c_1 \rightarrow c_2$). For each treadmill speed ($v$), these changes are computed as:
$${_{c_1}^{c_2}}\Delta\tau_v^{tot}=\tau_{c_2,v}^{tot}-\tau_{c_1,v}^{tot}$$ $${_{c_1}^{c_2}}\Delta\mu_v^{tot}=\mu_{c_2,v}^{tot}-\mu_{c_1,v}^{tot}.$$
After normalizing $\Delta \tau$ and $\Delta \mu$ across participants and walking speeds, we investigate their variation with respect to each other. \autoref{fig1} illustrates the possible outcomes:
\begin{itemize}
\item \textbf{Disagreement Increase (}$\bm{\Delta \tau>0}$ , $\bm{\Delta{\mu}>0}$\textbf{)}\\
This condition is associated with an increase in both the total interaction torque and the total muscular effort, indicating that switching from controller $c_1$   to controller $c_2$ has led to an elevation of human muscular effort. Consequently, their contribution to motion has increased. This, however, has resulted in a higher total interaction torque with the exoskeleton, implying that the applied torques by the exoskeleton are not aligned with the user's desired motion. As a result, the user needs to exert additional effort to correct the motion while contending against the applied torques from the exoskeleton. Thus, the increased interaction indicates a lack of harmony between the user's intentions and the assistance delivered by the exoskeleton.
\item\textbf{ Disagreement Decrease (}$\bm{\Delta \tau<0}$ , $\bm{\Delta{\mu}<0}$\textbf{)}\\ 
If controller $c_2$ demonstrates improved consistency compared to $c_1$ with the user's desired motion, the user will experience less resistance from the exoskeleton. This reduced discordance between the exoskeleton and the user's intended movements results in decreased total muscular effort and overall exertion by the user. Additionally, achieving a further reduction in disagreement between the human and exoskeleton can lead to one of the two next scenarios.
\item \textbf{Human Yields Control to Robot (}$\bm{\Delta \tau>0}$ , $\bm{\Delta{\mu}<0}$\textbf{)}\\
With the continued reduction in human-user interaction, a user may consider relinquishing motion control to the exoskeleton. This implies that the user will no longer actively contribute to the movement and will essentially deactivate their muscles, resulting in a decrease in overall muscular effort. Consequently, the exoskeleton must generate the necessary torque to facilitate the movement of both the exoskeleton and the passive dynamics of the human body. The increase in human-exoskeleton interaction torques, in this scenario, is not due to conflicts between the exoskeleton's motion and the user's desired motion but rather because the exoskeleton is effectively carrying the user's body.
\item \textbf{Human Takes Control (}$\bm{\Delta \tau<0}$ , $\bm{\Delta{\mu}>0}$\textbf{)}\\
In a contrasting scenario, the user may actively participate in the motion, resulting in an increased level of muscular effort. Consequently, the total interaction torque between the user and the exoskeleton may decrease. This reduction occurs because the human and exoskeleton motions are synchronized in time and consistent in space, creating a harmonious alignment between the two.
\end{itemize}

We conducted the aforementioned analysis at various speeds for the TBC→HTC, TBC$\rightarrow$AMTC, and HTC$\rightarrow$AMTC cases. We also performed a stride-wise analysis for the TBC$\rightarrow$HTC and TBC$\rightarrow$AMTC scenarios, which involves computing changes of total interaction force and muscle effort at each stride during the HTC and AMTC controllers as 
$$ {_{TBC}}^c\Delta\mu_{v,s} =\mu_{c,v,s} -\mu_{TBC,v}^{tot}$$ $$ {_{TBC}}^c\Delta\tau_{v,s} =\tau_{c,v,s} -\tau_{TBC,v}^{tot}  , $$
where $c\in \{\text{HTC}, \ \text{MBTC}\}$. These calculations allowed us to analyze the precise changes in interaction force and muscular effort for each stride during the HTC and AMTC controllers in relation to the TBC controller.

\subsubsection{Statistical Analysis}

To identify statistical differences, we initially employed a Friedman test with a significance level of 0.05 to test group-level differences. Following the Friedman test, we conducted pairwise comparisons, between the blocks, using the Wilcoxon signed-rank test. To account for multiple comparisons between the three blocks, we applied the Bonferroni correction.

\section*{Declarations}

\begin{itemize}
\item \textbf{Funding}\\
NSERC Discovery under Grant RGPIN-2018-4850, the Tri-Agency Institutional Programs Secretariat (TIPS): New Frontiers in Research Fund (NFRF)–Exploration under Grants 2018-1698, 2022-620, John R. Evans Leaders Fund Canadian Foundation for Innovation, Ontario Research Fund (ORF). 
\item \textbf{Conflict of interest/Competing interests }\\ No competing or conflict of interests.
\item \textbf{Code availability }\\
Contact A.A. for code and other materials. 
\item \textbf{Author contribution}\\ M.S. and A.A. designed the experiment. M.S. designed the AMTC and TBC controllers and implemented them on the Indego exoskeleton along with the HTC controller. M.S. and J.F. performed the experiment. M.S. analyzed and visualized the data. M.S. and A.A. interpreted the results and prepared the manuscript. All authors provided critical feedback on the manuscript.
\end{itemize}

\begin{appendices}

\section{Complementary Example Data}\label{secA1}

\begin{figure*}[h!]
	\centering	
	\scalebox{.9}{\includegraphics[trim=40 70 0 0]{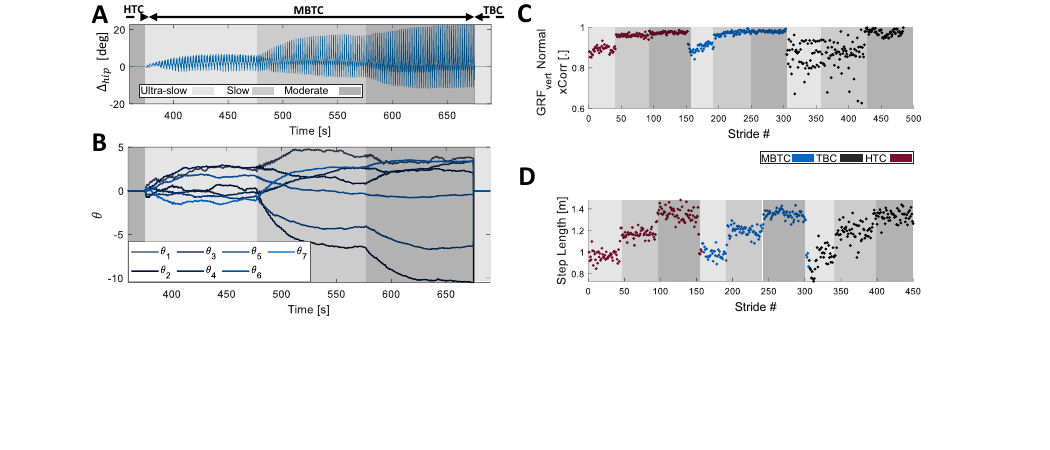}}
	\caption{(\textbf{A}) Evolution of the modification term for Participant \#1's right hip trajectory ($\Delta_{\text{hip}}$). $\Delta_{\text{hip}}$ is zero before and after the MBTC block. During the MBTC block, the modification term of the trajectory adapts to make the reference trajectory closer to the user joint angle. At each of the ultra-slow, slow, and moderate walking speeds, the modification pattern in the steady state converges to a different amplitude and shape, indicating that the user joint angle, and as a result, the exoskeleton reference trajectory has evolved to a different pattern at each speed. (\textbf{B}) Modification trajectory coefficients evolve during the MBTC block. Each coefficient has converged to a steady state value at the end of each treadmill speed condition. After the change in treadmill speed, coefficients converge to new optimum levels. These levels, as mentioned for the modification term of the hip trajectory, differ for each speed. (\textbf{C}) Pearson correlation between the GRF of each stride with the average GRF profile recorded during natural walking without the exoskeleton. (\textbf{D}) Stride length computed using the Physiolog 6s IMU sensors.}
	\label{fig11}
\end{figure*}

\myref{fig11}{A} illustrates the modifications made to the hip reference trajectory ($\Delta r_{\text{hip}}$) during the MBTC block. These modifications occur after the HTC block and before the TBC block, as depicted in (\autoref{fig10}), across ultra-slow, slow, and moderate walking speeds. Similarly, \myref{fig11}{B} showcases the adaptation of the Fourier coefficients ($\bm{\theta}$) and their convergence pattern during the MBTC block at ultra-slow walking. The Fourier coefficients begin to adapt as soon as that block starts. It takes approximately 20 seconds for them to converge, and they subsequently exhibit minor oscillations around their converged values until the treadmill speed is switched to the slow speed. This change in treadmill speed induces new walking conditions, resulting in distinct gait patterns. Consequently, the Fourier coefficients converge toward a new set of steady-state values. A similar pattern emerges when the treadmill is switched to the moderate speed. Pearson correlation between the vertical Ground Reaction Forces during each of the HTC, MBTC, and TBC blocks and natural walking with no exoskeleton is plotted for each stride in \myref{fig11}{C}. According to the graph, the GRF patterns resemble the GRF in natural walking at slow and moderate walking speeds. In ultra-slow speed, however, the correlation slightly decreases in all three blocks. Finally, \myref{fig11}{D} shows the computed stride length. As expected, in all three blocks, the stride length increases with the increase in treadmill speed.

\section{Comparison with Natural Walking}
To identify which controller led to a gait closer to natural walking, we analyzed the spatiotemporal parameters of gait (obtained using GaitUp Sensors) in TBC, HTC, and AMTC and compared them to those of each participant's natural walking with no exoskeleton. Moreover, we analyzed the correlation of recorded ground reaction torques in the case of each controller with natural walking.
\subsection{Gait Spatiotemporal Parameters}

\begin{figure*}[h!]
	\centering	
	\scalebox{1}{\includegraphics[trim=0 0 0 0]{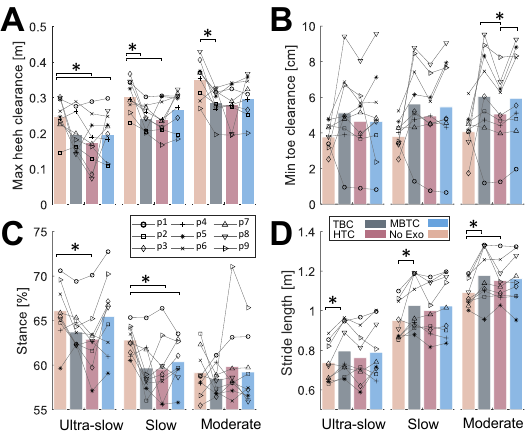}}
	\caption{The average maximum heel clearance (\textbf{A}), minimum toe clearance (\textbf{B}), stance percentage (\textbf{C}), and stride length (\textbf{D}), computed for each of the ultra-slow, slow, and moderate speeds during natural walking with no exoskeleton, TBC, HTC, and the MBTC blocks for each participant. Bars show the average of each metric across participants.}
	\label{fig12}
\end{figure*}

\myref{fig12}{A} shows the maximum heel clearance in natural walking condition (with no exoskeleton) and during exoskeleton walking at each of the TBC, HTC, and AMTC blocks at ultra-slow, slow, and moderate speeds. Natural walking has the greatest heel clearance, which is expected as the addition of the exoskeleton weight makes it harder for the participant to lift their foot to their convenient level. Among the three controllers, AMTC has a heel clearance closer to that of natural walking, indicating that maintaining the desired heel clearance is easier for the participant in the case of this controller. Similarly, \myref{fig12}{B} compares the minimum toe clearance, showing that at all walking speeds, natural walking with no exoskeleton has the lowest minimum toe clearance. The HTC controller demonstrates lower toe clearance than the TBC and AMTC at all speeds. However, these differences are not statistically significant unless during moderate walking, where the HTC has led to a 21.3\%±18.2 and 12.2\%±9.3 decrease in minimum toe clearance compared to the AMTC and TBC, respectively (Friedman: $p<$0.001, Wilcoxon signed rank: $p_{_{HTC,AMTC}}<$0.007, $p_{_{HTC,TBC}}<$0.003).

Regarding the stance percentage, as depicted in \myref{fig12}{C}, AMTC exhibits the closest behavior to natural walking at all speeds. In all cases, TBC has a smaller stance percentage compared to natural walking. During ultra-slow and slow-speed walking, HTC has the smallest stance percentage. In moderate-speed walking, however, HTC has the highest stance percentage. Finally, \myref{fig12}{D} illustrates the stride length for each of the controllers at the three experimented walking speeds. The stride length generally increases with the increase in treadmill speed. At each individual speed, however, natural walking appears to have the smallest stride length on average, mostly compared to the TBC, which has the highest stride length (Friedman: $p<$0.05, Wilcoxon signed rank: $p<$0.01 for all speeds).

\subsection{Ground Reaction Force}

\begin{figure*}[h!]
	\centering	
	\scalebox{1}{\includegraphics[trim=0 0 0 0]{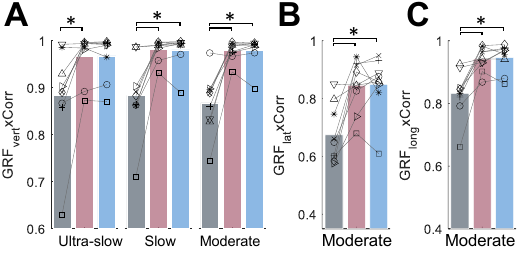}}
	\caption{(\textbf{A}) The average Pearson correlation computed between the average vertical GRF in natural walking without the exoskeleton and the vertical GRF during walking with each of the proposed controllers at different speeds. Depicted bars show the average of each metric across participants. (\textbf{B}) and (\textbf{C}) shows the similar graph for the lateral and longitudinal GRF plotted only during the moderate speed.}
	\label{fig13}
\end{figure*}

\myref{fig13}{A} shows the Pearson correlation of the vertical ground reaction force between walking with each of the controllers and natural walking without the exoskeleton. At all speeds, TBC exhibits the most different vertical GRF compared to the other controllers (Friedman: $p_{{\text{Ultra-slow}}}<$0.02, $p_{{\text{Slow}}}<$ 0.001, and $p_{{\text{Moderate}}}<0.016$; Wilcoxon signed rank: $p<$0.01 for all speeds). In contrast to the TBC, HTC and AMTC led to more natural vertical GRF profiles with Pearson correlations greater than 96\% at all speeds. The same observation stands for the GRFs in lateral (\myref{fig13}{B}) and longitudinal (\myref{fig13}{C}) directions, where TBC shows the least natural GRF profile while HTC and AMTC demonstrate closer to natural walking GRF with Pearson correlations greater than 82.2\% and 91.5\% in lateral and longitudinal directions, respectively (Friedman: $p<$0.003; Wilcoxon signed rank: $p<$0.004 for all cases).

\end{appendices}

\bibliography{GeneratedBibTex}

\end{document}